# Using Decision Tree as Local Interpretable Model in Autoencoder-based LIME

1st Niloofar Ranjbar
*Amirkabir University of Technology*
*Persian Gulf University*
nranjbar@aut.ac.ir
nranjbar@pgu.ac.ir

2nd Reza Safabakhsh
*Amirkabir University of Technology*
safa@aut.ac.ir

*Abstract*— Nowadays, deep neural networks are being used in many domains because of their high accuracy results. However, they are considered as "black box", means that they are not explainable for humans. On the other hand, in some tasks such as medical, economic, and self-driving cars, users want the model to be interpretable to decide if they can trust these results or not. In this work, we present a modified version of an autoencoder-based approach for local interpretability called ALIME. The ALIME itself is inspired by a famous method called Local Interpretable Model-agnostic Explanations (LIME). LIME generates a single instance level explanation by generating new data around the instance and training a local linear interpretable model. ALIME uses an autoencoder to weigh the new data around the sample. Nevertheless, the ALIME uses a linear model as the interpretable model to be trained locally, just like the LIME. This work proposes a new approach, which uses a decision tree instead of the linear model, as the interpretable model. We evaluate the proposed model in case of stability, local fidelity, and interpretability on different datasets. Compared to ALIME, the experiments show significant results on stability and local fidelity and improved results on interpretability.

## I. Introduction

Deep neural networks have shown remarkable accuracy in many machine learning tasks, while they are considered as "black box" models due to their weak interpretability. It is hard for humans to understand the internal structure of these networks. In some critical tasks such as disease diagnosis, users need to understand the model well and know why the model produced some output for a particular input. In many tasks, users want to gain knowledge from the model and ensure if the results are correct, safe, and unbiased. Furthermore, developers can identify model errors and fix them easily when they understand the model [1]. Explainable AI can be divided into several categories in terms of scope, methodology, and usage [2]. The scope of explanations produced by an algorithm can be local or global. Locally explainable methods explain the individual features of a single instance of input data from the data set while globally explainable models explain the decision of the model as a whole. On the other hand, the methodology of algorithm implementation can be based on backpropagation or data-perturbation. In the first method the feature importance is calculated when gradients propagate from the last layer into the first layer. In contrast, the second method perturbs a given input instance's feature set by sampling data randomly near the instance or using data augmentation approaches.

An explainable method can make a neural network explainable intrinsically, or it can be applied as an external algorithm to produce explanations. Some approaches try to change the architecture of the model itself to make it interpretable, which falls into the model-intrinsic category [3] [4] [5]. Others are model-agnostic or post-hoc algorithms that can explain the predictions of already existing neural networks without the need to train them again [6] [7] .

LIME [8] is a post-hoc approach that tries to generate explanations locally for an instance of interest by perturbing random samples near the instance and then training a linear model locally on this new data set. Then, the weights of the linear model are reported as the explanation. Many researchers have tried to improve LIME from different viewpoints [9] [10]. For example, the DLIME method [11] uses hierarchical clustering (HC) to group training data and then uses the cluster closest to the instance to train interpretable models locally. By these modifications, they tried to solve the lack of "stability" problem. Another modification of LIME is ALIME [12] which uses an autoencoder-based weighting function to solve the lack of fidelity and stability in LIME. Although all of these methods use linear models to produce explanations, none of them have evaluated the quality and usability of the produced explanations from the human point of view. The primary purpose of these models is to produce explanations that are understandable for humans. In addition, the method's performance has not been evaluated when other interpretable models such as the decision tree are used instead of the linear model. This paper introduces a new version of ALIME named tree-ALIME that uses a decision tree instead of a linear model as a locally interpretable model. We evaluate both ALIME and tree-ALIME in the case of stability, fidelity, and interpretability from the human's point of view on four tabular data sets and sentiment analysis tasks. All of the source codes and links are available in our GitHub repository (The link is at the end of Paper) . Our contributions are as follows:

- Introducing a new version of ALIME, which uses a decision tree instead of a linear model.

- Introducing a new measure for evaluating the stability of models.

- Evaluating interpretability quantitatively and from the human's point of view for both ALIME and tree-ALIME on four different tabular data sets and text classification task.

The rest of the paper is organized as follows: First, in Section II LIME and ALIME algorithms are explained in detail. Then in Section III the tree-ALIME algorithm is introduced. Finally, in Section IV evaluations details and results are presented.

## II. Backgraound

### A. LIME

Local interpretable model-agnostic explanations (LIME) is a famous work introduced recently. This algorithm first samples new data randomly around an instance of interest.

After that, it weighs the samples according to their similarity to that instance using some weighting function, such as Euclidean distance. Then, the algorithm uses the black-box model to predict labels for the newly generated data set and uses them as the ground truth. Next, the algorithm trains an interpretable model such as a linear model on these weighted generated points. Finally, the prediction is given by the coefficients of the local linear model.

This learned model should be a good approximation of the black-box model locally, which means it should have local fidelity. Besides, the explanations generated, for instance, in different runs should not be changed, which means the algorithm should have stability. One of the problems of LIME is the lack of local fidelity and stability. One of the algorithms introduced to solve this problem is the autoencoder-based LIME (ALIME).

*B. ALIME*

ALIME is a version of LIME that trains a denoising autoencoder and uses it as a weighting function for the generated data around an instance. First, it computes the latent embeddings for the generated data set and the desired point and then measures the distance on the embedded space. After that, it uses an exponential kernel as a function of distance to weigh the points. Both LIME and ALIME use a Gaussian distribution to generate new data. However, there is a difference between them. ALIME generates many points randomly and then gathers the n closet points to an instance as the new data set, while LIME generates n random points around an instance of interest directly. The ALIME algorithm is described in Algorithm 1 [12].

However, both LIME and ALIME use a linear model as the interpretable model to be trained locally. In the next section, we present a new version of ALIME called tree-ALIME, which uses a decision tree instead of a linear model as the interpretable model. Besides, we present a new approach to test the LIME itself on the text classification task.

---

**Algorithm 1:** ALIME

**Input:** Dataset $D_{train}$ with K features, Trained Model M , instance x , number of points used n , sampled points m
**Output:** feature importance at x

/* compute embeddings using autoencoder*/

begin

1: $AE \leftarrow AutoEncoder.fit(D_{train})$ ; // train Autoencoder

2: $D_{sample} \leftarrow Gaussian(K)$ ;

3: $E \leftarrow [AE(y)|y \in D_{sample}]$ ;

4: $E(x) \leftarrow AE(x)$

end

/* Given x, calculate feature importance*/

begin

1: $d \leftarrow |E - E(x)|$ ;

2: $d_{min} \leftarrow n\_th\_min(d)$ ;

3: find y such that $[ |E(y) - E(x)| < d_{min} ]$ ;

4: $W \leftarrow [e^{-z}|z \in d]$ ; // Calculate weights

5: $L \leftarrow LinearModel.fit(local\_data, W)$ ; // Fit linear model

6: return $L_w$ ; // Return weights of the linear model

end

---

## III. TREE-BASED ALIME

In this section, we first present the proposed algorithm, treeALIME, in detail and then propose a new approach for using the LIME and tree-ALIME algorithms in the text classification task.

*A. .Tabular data*

The algorithm which we proposed for Tabular data is same as ALIME with one basic difference. We used a decision tree instead of a linear model as the interpretable model. The proposed algorithm steps are as follows:

**Preprocessing:** First, we change all the categorical features to binary by one-hot encoding. After that, we scale all instances simply with sklearn.preprocessing.scale package of sklearn library in python. It centers all the points to the mean and changes them to have unit variance after scaling.

**Training the black-box model:** After preprocessing the data, we split the data into train and test sets randomly. We choose 80% for training and 20% for testing. After that, we choose the black-box model and train it with the normalized training set. Here we used a feed-forward neural network with two hidden layers. The the number of neurons in each layer is tuned for each dataset separately.

**Training a denoising autoencoder**: We use an already implemented denoising autoencoder and train it with the normalized data.

**Perturbing data:** For every dataset, we use the Gaussian distribution for generating 10000 random samples. We use mean=1 and variance=0 for categorical features; but for others, we use the mean and variance of the training dataset, just same as the LIME. After that, we normalize all the newly generated data as we do for the main dataset.

**tree-ALIME:** At the final step, we perform ALIME-tree as presented in Algorithm 1 with one main difference, which is fitting a decision tree instead of the linear model as the local model. The Decision Tree we used here is CART.

*B. Text Classification Task*

We use the sentiment analysis task as a classification task here. The proposed algorithm steps for this section are as follows:

**Preprocessing**: First, we clean the text by removing links, email addresses, tabs, and all of the symbols that are not alphabetic symbols or numbers or one of these symbols: "!" , "? ". After that, we tokenized all of the reviews by the XLNet Tokenizer.

**Training the black-box model**: We use the XLNet model as the black-box model here. As the XLNet [13] has the best results in the sentiment analysis task on the IMDb dataset, and we use this dataset for our evaluations,and XLNet as the black-box model. We used the source code implemented by Shanayghag for training XLNet on the IMDb dataset (The source code link is provided at the end of the paper). In this source code the transformers API is used.

**Perturbing data**: We used the "nlpaug" library to generate new data. With this library, one can use different ways to augment the data, and in this case, the text. . We used several augmenters as follows:

- **Contextual Word Embeddings** Augmenter which uses BERT [14], DistilBERT [15], RoBERTA [16],

or XLNet as the contextual word embedding to insert a new word between other words or change some words in the text.

- **Synonym** Augmenter which uses WordNet or ppdb to replace the words with their synonyms.
- **Antonym** Augmenter which uses WordNet to replace the words with their antonyms. We use this augmenter because we need some review classes to be changed.
- **Random Word** Augmenter which swaps or removes some words from the text.

After generating new reviews, we choose n closest ones to the review, for which we want to generate an explanation. The similarity between the reviews is measured as described in the next step.

**Weighting Reviews:** To weigh the reviews, we first, use a sentence-transformer model called "Sentence-BERT" [17] to map the text to a 768-dimensional dense vector space and then calculate cosine similarity between the encoded texts. Finally, we use the weighting function as introduced in Algorithm 1 to weigh the generated texts according to their similarity to the review of interest.

**Feature Extraction:** After gathering the local dataset, we need to extract features from the text and train the interpretable model. First, we preprocess the generated texts by changing all characters to lower case and removing punctuation marks and stop words. After that, we use the TfidfVectorizer from "sklearn.feature_extraction" library to calculate the TF-IDF score for all of the words and use them as features. Besides, we use "sentiwordnet" to find the sentiment of all of the words in the text and count the number of positive and negative words as two extra features. Note that these two features are used when we choose a decision tree as the interpretable model.

**Training the Local Model:** In this phase, we use Algorithm 1 to train an interpretable model locally. In the experiments, we refer to the linear model as "ALIME" and the decision tree model as "tree-ALIME".

IV. EXPERIMENTS AND RESULTS

In this section, we first introduce the datasets we used for the experiments. Then, we discuss the models' parameter settings. After that, we provide the results of evaluating models from three points of view: local fidelity, stability, and interpretability.

*A. Datasets*

We use four tabular datasets Breast Cancer [18], Hepatitis [19], Indian Liver Patient [20] and Adults [21] and one textual dataset for the experiments. Table I shows the number of features and instances in the tabular datasets. We use the IMDb dataset [22] for binary sentiment classification which contains 49582 movie reviews.

TABLE I. THE DATASETS DETAILS

| Dataset Name | Features | Instances |
|---|---|---|
| Breast Cancer | 30 | 569 |
| Hepatitis | 19 | 155 |
| Indian Liver Patient | 10 | 583 |
| Adults | 14 | 48842 |

TABLE II. PARAMETERS AND ACCURACY OF TABULAR DATASETS

| Dataset Name | layer1 | layer2 | Accuracy |
|---|---|---|---|
| Breast Cancer | 20 | 25 | 97.5% |
| Hepatitis | 25 | 30 | 81.57% |
| Indian Liver Patient | 5 | 5 | 71.35% |
| Adults | 20 | 5 | 85% |

*B. Parameter setting*

1) Black-box models parameters: To tune the feed-forward neural network parameters and find the accuracy of models, K-Fold cross-validation with K=10 is used. In every iteration of K-Fold, 90% and 10% of the training data is used for training and validation, respectively. After that, the Grid Search, is used to find the best number of neurons for each layer. The number of neurons is changed in the range of 5 to 35 with step 5.

The iteration which has the most accuracy on the test-set is chosen as the best split, and the number of neurons found in that by the grid search is chosen as the best number of neurons. The average of accuracies on the test-set in all iterations is calculated and chosen as the final accuracy. All the models are trained at a maximum of 150 epochs with an early stopping condition on the loss of validation data. The Adam Optimizer and binary cross entropy loss are used for all of the models. For the IMDb dataset, 50% of the data is used as the training set, 25% as the validation set, and 25% as the test set. The XLNet model is trained for three iterations. Its predictive power is tested on the test set, and gained 94.8% accuracy. The best parameters and accuracies on the test set for each dataset are shown in Table II.

2) Interpretable models parameters: The number of randomly generated points for all of the tabular datasets is chosen as m=10000. For the IMDb dataset, n=100 similar reviews to the review of interest are generated with data augmentation methods as described in Section III-B. As the source code of ALIME is not available publicly we implemented it. We use the Logistic Regression model as the linear model and train it for a maximum of 150 iterations. The decision tree is used as an interpretable model in our tree-ALIME algorithm. Note that we used Colab GPUs for both training and evaluation sections.

*B. Results*

1) **Local-Fidelity:** To evaluate the local fidelity, we first generate new points around every data point in the test set. Then we train the interpretable model using the generated data points. After that, the data point is given as an input to the trained model. The predicted class for the data point is then compared with the predicted class by the black-box model. At last, the number of data points in the test set for which these two predicted classes are equal is divided by the number of data points in the test set, and the accuracy is calculated as the local fidelity measure.

As both of the interpretable models produce the prediction probability of each class as output, the probabilities are changed into class labels by using a threshold equal to 0.5. If the probability is less than the threshold, the class is chosen as 0; otherwise, it is chosen as 1. To find out the effect of using a low or high number of samples around an instance of interest, we use the three tabular datasets "breast-cancer", "hepatitis" and "IndianLiverPatientDataset(ILPD)", and calculate the accuracy as the local fidelity measure for all of them. The results for the three datasets are shown in Figures

1, 2 and 3, respectively. These figures show that, for the three datasets, when the number of samples increases, the accuracy of the linear model is increasing too. Besides, for all of them the accuracy of the linear model is more than the decision-tree model, which means the decision-tree model has a lower local fidelity.

For the breast cancer dataset (Fig.1), the number of samples that cause the best local fidelity for the linear model is more than 1200, while for the decision tree, it is not dependent on the number of samples and does not have a regular pattern. The best local fidelity for the decision tree on this dataset is 0.9298.

On the other hand, for the hepatitis dataset (Fig. 2), the linear model and the decision tree model have comparable results at a low number of samples. The best local fidelity, which equals 1.0, happens at n=800 and n=900 for the linear model and the decision tree, respectively. However, as the number of samples increases, the local fidelity decreases for the decision tree model and increases for the linear model.

For the liver patient dataset (Fig. 3), the overall local fidelity is less, and the differences between the local fidelity of the models are more than the other two datasets. The linear model has a higher local fidelity, which increases with the number of samples. The first time it gains the best local fidelity, (0.9487), is when the number of samples is 1900. However, the decision tree model's best result (0.8205), happens when the number of samples is 2000. The difference between the best results of the models here is 0.1282, while for the hepatitis dataset, it is zero, and for the breast cancer dataset, it is 0.0702. For the adult dataset, we did not vary the number of samples. We set the number of samples equal to 1000. Besides, we used only 1000 data points of the test dataset to do the test. For the linear model, the local fidelity is 0.794, and for the decision tree model, it is 0.789. This shows no significant difference between them, and they are close to each other.

For the text classification task, we randomly choose 100 reviews from the test set. The linear model's local fidelity is 0.93, and the decision tree model's is 0.92. As can be seen for both of these models, the local fidelity is acceptable, but the decision tree's fidelity is slightly less than the linear model's fidelity.

We conclude that the local fidelity of the linear model is higher than the decision tree model, but there is not a significant difference between them. The probable cause is the simple version of the decision tree we used. Other versions of the decision tree have better accuracy, and using them may cause better results. Another possible reason is that the decision tree tries to define an exact split point for each node which may cause it to overfit the training data.

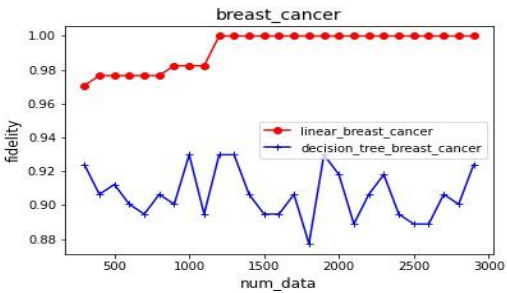

Fig. 1: local fidelity of the breast cancer dataset

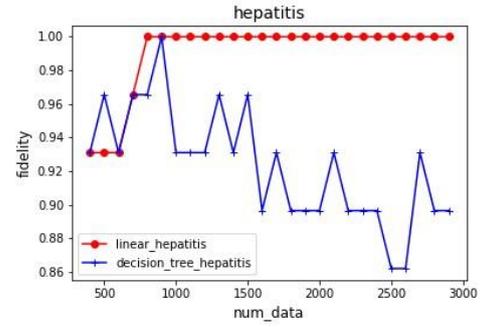

Fig. 2: local fidelity of the hepatitis dataset

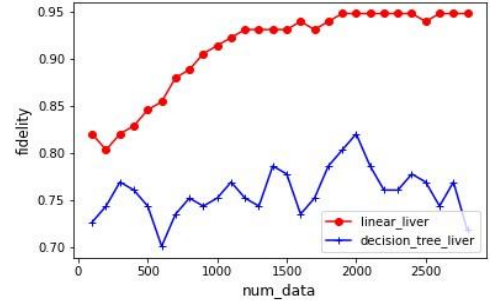

Fig. 3: local fidelity of the liver patient dataset

2) **Stability:** As mentioned before, the produced explanation for an instance of interest must not change in different runs. That is the definition of stability. The stability of the models is calculated as follows.

First, a data point from the test set is chosen randomly. Then each algorithm runs 20 times with 20 different random seeds, which creates different perturbed data points around the randomly chosen test data in each run. For the linear model, we want to determine whether the same features have the most impact on the results in each run. So we choose 25% of the highest positive coefficients and 25% of the lowest negative coefficients as the coefficients which have the most effect. After that, we calculate the stability score for the positive and negative coefficients of the linear model separately with equations (1) and (2) and then calculate the average of them as the final stability score for the linear model (equation (3)):

$$linear\_p\_stability = \frac{\sum_{i=1}^{P}\sum_{j=1, i\neq j}^{P}\frac{length(p_i \cap p_j)}{length(p_i \cup p_j)}}{P(P-1)} \quad (1)$$

$$linear\_n\_stability = \frac{\sum_{i=1}^{N}\sum_{j=1, i\neq j}^{N}\frac{length(n_i \cap n_j)}{length(n_i \cup n_j)}}{N(N-1)} \quad (2)$$

$$linear\_stability = \frac{linear\_p\_stability + linear\_n\_stability}{2} \quad (3)$$

Where P=N=20 is the number of runs. $p_i$ is the positive coefficients set in the run number i, and $n_i$ is the negative coefficients set in the run number i, and the "length( )" function returns the set length.

For the decision tree model, we want to determine whether the same features are used for building the decision tree in each run or not. So the features with importance scores of larger than zero are selected in each run, and the stability score is calculated with equation (4):

$$tree\_stability = \frac{\sum_{i=1}^{T}\sum_{j=1,i\neq j}^{T}\frac{length(t_i \cap t_j)}{length(t_i \cup t_j)}}{T(P-1)} \quad (4)$$

Where T=20 is the number of runs, $t_i$ is the set of features used in the run number i. The closer the stability score is to 1, the more stable it is; and the closer it is to zero, the less stable it is. We vary the number of perturbed samples as in calculating the local fidelity and calculate the stability for both methods and for all of the tabular datasets. The results are shown in the Fig. (4)- (7).

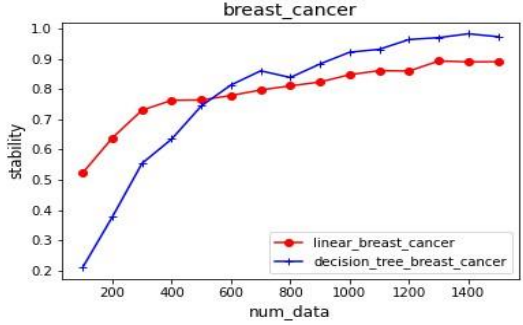

Fig. 4: stability of the breast cancer dataset

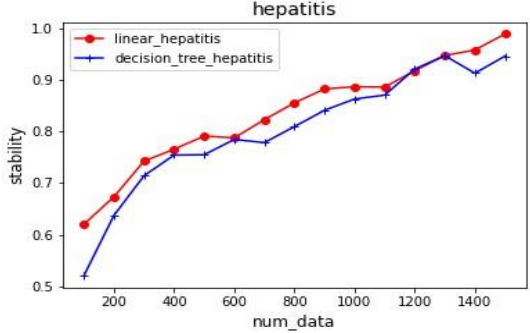

Fig. 5: stability of the hepatitis dataset

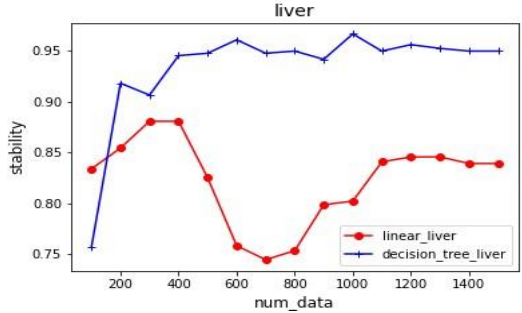

Fig. 6: stability of the liver patient dataset

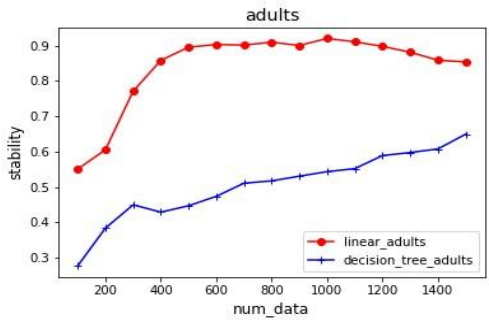

Fig. 7: stability of the adults dataset

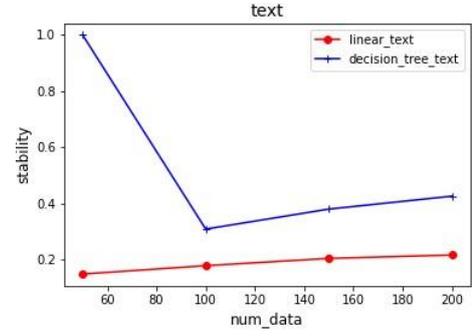

Fig. 8: stability of the IMDb dataset

As can be seen in Fig. 4, for the breast cancer dataset, the stability of the linear model is initially much more than that of the decision tree model. However, as the number of data points used for training increases, the stability of the decision tree also increases. This is because the decision tree tends to overfit as the number of training data grows up. As a result, the number of features used in the decision tree increases, so the stability score also increases. However, this does not happen for the linear model because we use a fixed number of features to calculate the stability in each run. Thus the number of features does not affect the stability of the linear model. However, the average number of features used in the linear model and the decision tree model for this dataset is 12.33 and 25.15, respectively. This shows that the decision tree uses most of the features, which is not interesting.

Likewise, for the hepatitis dataset (Fig. 5), the stability of both the linear method and the decision tree method increases with the number of data used for their training. However, for this dataset, the average number of features used in the linear and decision tree models is 9 and 14.3, respectively. As can be seen, the results are better for this dataset, and the difference between the number of used features is less than the breast cancer dataset.

On the other hand, for the liver dataset (Fig. 6), the decision tree model has a higher stability score than the linear model, even when the number of data used for training these models is low. However, there is a drop in the stability score of the linear model when the number of data increases from 400 to 600. The stability score almost remains the same for the decision tree from the number of data equal to or greater than 600. However, the average number of features used in the linear and decision tree models is 5 and 9.29, respectively. So, with only four more features, the decision tree can achieve a higher stability scores than the linear model for this dataset.

Furthermore, for the adults dataset (Fig. 7), the linear model has a much higher stability score than the decision tree. However, the average number of features used in the linear and decision tree models is 24.13 and 16.9, respectively. As can be seen, here, the average number of features used by the decision tree is less than the linear model, and consequently, its stability score is more diminutive. As the dataset has 105 features, we can ignore the difference between the features used for these models, while there is a big difference between their stability scores.

In contrast, for the IMDb dataset (Fig. 8), the decision tree model has a higher stability score than the linear model. The number of features used in the linear and decision tree models is 146 and 8.86 for this dataset. These results show that while the decision tree does not use many features, its stability score is significantly higher than that of the linear model.

To sum up, it seems that the stability depends on the number of features used by the algorithm and the results are dependent on the dataset. As such, one cannot declare a method superior to the other one.

3) **Interpretability:** To evaluate the interpretability of the models, we used crowdsourcing. We asked evaluators who were familiar with machine learning models to evaluate the models. For every dataset, we choose 20 random data points from the test set. We choose the five highest positive coefficients and five lowest negative coefficients as the most effective coefficients for the linear model and generate explanations for the linear model. For the decision tree model, we choose a maximum depth of 5 and present the decision tree as the explanation to the evaluators. Examples of explanations generated for the linear model and the decision tree model for the hepatitis dataset are shown in the appendix section. More examples are available in our GitHub repository. We evaluate the interpretability in two ways: First, examiners are asked to rate the clarity of the explanations from 1 to 5, where 1 means very unclear and 5 means very clear. The averaged scores of examiners for all of the datasets are shown in Fig. (9) - (13). As seen for all datasets, the decision tree model has a significantly higher clarity score than the linear model. Especially for the breast cancer, hepatitis and adults dataset, this is more remarkable.

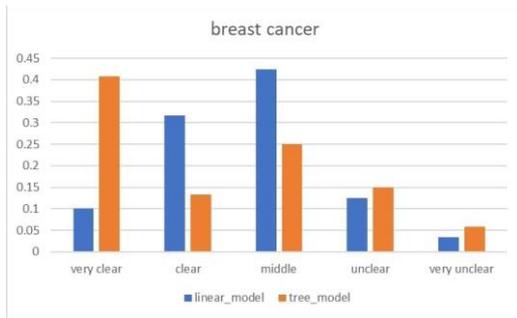

Fig. 9: Interpretability scores of breast cancer

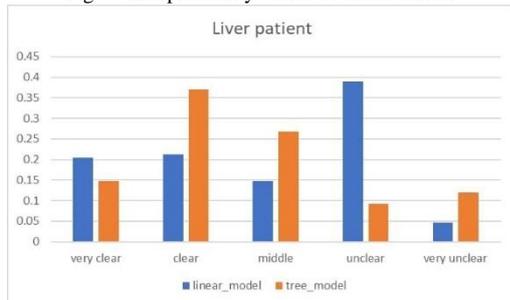

Fig. 10: Interpretability scores of liver patients

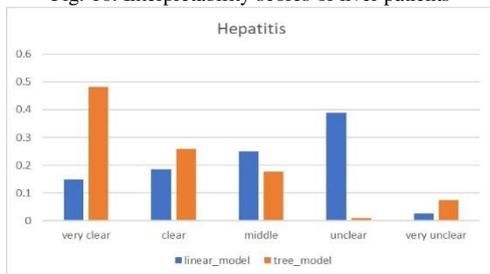

Fig. 11: Interpretability scores of hepatitis

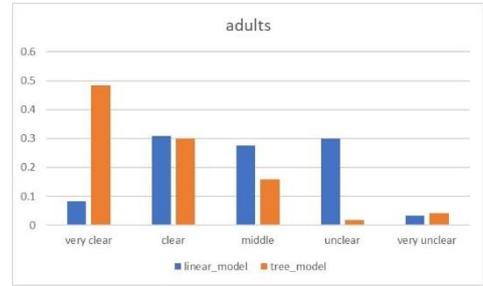

Fig. 12: Interpretability scores of adults

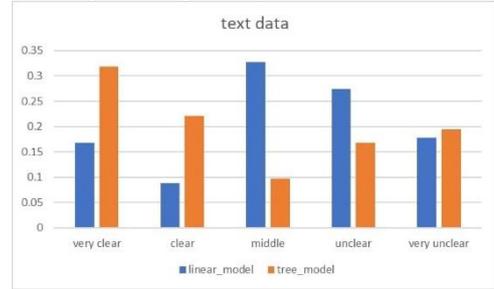

Fig. 13: Interpretability scores of IMDb

TABLE III. ACCURACY OF EXAMINERS PREDICTIONS

| Dataset | linear | decision tree |
|---|---|---|
| **Breast Cancer** | 0.37 | 0.62 |
| **Indian Liver Patient** | 0.72 | 0.52 |
| **Hepatitis** | 0.37 | 0.72 |
| **Adults** | 0.51 | 0.85 |
| **IMDb** | 0.5 | 0.72 |

Second, to investigate the usefulness of explanations, we asked the examiners to use the explanations and the information of data points and guess the label of each data point. After that, we calculated the accuracy of predicted labels. Results for all of the datasets are shown in Table III. Note that examiners did not have access to the true labels of data points. Examiners could make correct predictions for all of the datasets except the liver patients dataset. For the linear model, examiners could not guess the label even better than random for some datasets, which shows the linear model explanations are not understandable enough for users. However, for the decision tree model, things are different. Examiners could always predict the labels better than random, and for four datasets, accuracy is considerably high.

V. CONCLUSION

This paper introduced a new version of ALIME, which uses the decision tree model instead of the linear model as the locally interpretable model. We evaluated our method on four tabular datasets and the IMDB dataset regarding local fidelity, stability, and human point of view interpretability. Although the linear model outperforms the decision tree model in terms of local fidelity, its stability was almost equal to the decision tree model. However, the decision tree model gains significantly better results in terms of interpretability, which is the essential factor in the explainable models.

For future work, one should consider using other versions of the decision tree such as random forest, which is better than the simple decision tree in terms of accuracy. In addition, other methods can be considered to produce new data points around a sample of interest.

## VI. APPENDIX

Examples of explanations generated for the linear model and the decision tree model for the hepatitis dataset are shown in the Figure14 and Figure15.

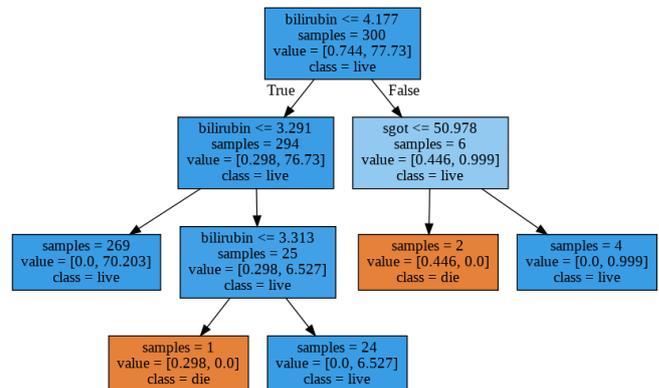

Fig14 : An example of explanation generated by the decision tree for a data from the hepatitis dataset

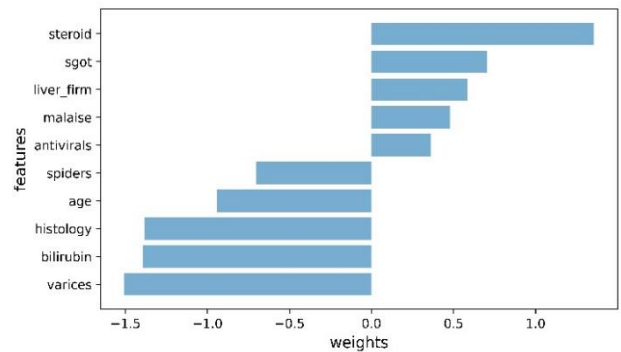

Fig. 15: An example of explanation generated by the linear model for a data from the hepatitis dataset

All of the codes and APIs links are as follows:

- https://www.kaggle.com/rspadim/simple-denoise-autoencoder-with-keras
- https://huggingface.co/transformers/model_doc/xlnet.html#xlnettokenizer
- https://github.com/shanayghag/Sentiment-classification-using-XLNet/blob/master/Sentiment_Analysis_Series_part_1.ipynb
- https://huggingface.co/transformers/model_doc/xlnet. html
- https://nlpaug.readthedocs.io/en/latest/#nlpaug
- https://scikit-learn.org/stable/modules/generated/sklearn. feature_extraction.text.TfidfVectorizer.html
- https://www.nltk.org/_modules/nltk/corpus/reader/ sentiwordnet.html
- https://scikitlearn.org/stable/modules/generated/sklearn.linear _model.LogisticRegression.html
- Our Github repository: https://github.com/nranjbar/interpretable_machine_learning